\documentclass[11pt,a4paper]{article}
\usepackage[hyperref]{acl2021}
\usepackage{times}
\usepackage{latexsym}

\usepackage{microtype}
\usepackage{graphicx}
\usepackage{booktabs}  
\usepackage{multirow}
\usepackage{subfigure}
\usepackage{verbatim}
\usepackage{CJKutf8} 
\usepackage{flushend}
\usepackage{makecell}
\usepackage{amsmath}
\usepackage{amssymb}
\usepackage{placeins}
\usepackage{float}
\aclfinalcopy % Uncomment this line for the final submission
\title{Fastformer: Additive Attention Can Be All You Need}

\author{Chuhan Wu$^\dagger$~~~~Fangzhao Wu$^\ddagger$~~~~Tao Qi$^\dagger$~~~~Yongfeng Huang$^\dagger$~~~~Xing Xie$^\ddagger$\\
    $^\dagger$Department of Electronic Engineering \& BNRist, Tsinghua University, Beijing 100084, China  \\
     $^\ddagger$Microsoft Research Asia, Beijing 100080, China\\
  {\tt\{wuchuhan15, wufangzhao, taoqi.qt\}@gmail.com} \\ {\tt yfhuang@tsinghua.edu.cn, xingx@microsoft.com}
  }
\date{}

\begin{document}
\maketitle

\begin{abstract}

Transformer is a powerful model for text understanding.
However, it is inefficient due to its quadratic complexity to input sequence length.
Although there are many methods on Transformer acceleration, they are still either  inefficient on long sequences or not effective enough.
In this paper, we propose \textit{Fastformer}, which is an efficient Transformer model based on additive attention.
In \textit{Fastformer}, instead of modeling the pair-wise intractions between tokens, we first use additive attention mechanism to model global contexts, and then further transform each token representation based on its interaction with  global context representations.
In this way, \textit{Fastformer} can achieve effective context modeling with linear complexity.
Extensive experiments on five datasets show that \textit{Fastformer} is much more efficient than many existing Transformer models and can meanwhile achieve comparable or even better long text modeling performance.

\end{abstract}

\section{Introduction}

Transformer~\cite{vaswani2017attention} and their variants have achieved great success in many fields. For example, Transformer is the backbone architecture of many state-of-the-art pre-trained language models in NLP, such as BERT~\cite{devlin2019bert} and GPT~\cite{radford2019language}.
Transformer also shows great promises in vision-related tasks~\cite{dosovitskiy2020image}.
The core of a Transformer model is self-attention mechanism, which allows the Transformer to model the contexts within an input sequence~\cite{parikh2016decomposable}.
However, since self-attention computes the dot-product between the input representations at each pair of positions, its complexity is quadratic to the input sequence length~\cite{vaswani2017attention}.
Thus, it is difficult for standard Transformer models to efficiently handle long input sequences~\cite{tay2020efficient}.

There are many methods to accelerate the Transformer model~\cite{beltagy2020longformer,zaheer2020big,wang2020linformer,Kitaev2020reformer,tay2021synthesizer}.
For example, BigBird~\cite{zaheer2020big} computes sparse attention instead of a dense one. 
It uses a combination of local attention, global attention at certain positions and random attention between a certain number of tokens.
However, sparse attention usually cannot fully model the global context~\cite{wu2021hi}.
Linformer~\cite{wang2020linformer} exploits the low-rank characteristic of the self-attention matrix by computing approximated ones.
It projects attention key and value into  low-dimensional matrices that are independent of the sequence length.
However, the approximation is in fact context-agnostic, which may weaken the context modeling ability of Transformer.
In addition, these methods are not efficient enough when the input sequence length is very long.

In this paper we propose \textit{Fastformer}\footnote{A pytorch version of \textit{Fastformer} using the huggingface style is available at https://github.com/wuch15/Fastformer.}, which is an efficient Transformer variant based on additive attention that can achieve effective context modeling in linear complexity. 
In \textit{Fastformer}, we first use additive attention mechanism to summarize the input attention query matrix into a global query vector.
Next, we model the interaction between attention key and the global query vector via element-wise product to learn global context-aware key matrix, and we further summarize it into a global key vector via additive attention.
Then we use element-wise product to aggregate the global key and attention value, which are further processed by a linear transformation to compute the global context-aware attention value.
Finally, we add together the original attention query and the global context-aware attention value to form the final output.
We conduct extensive experiments on five benchmark datasets in various tasks, including sentiment classification, topic prediction, news recommendation and text summarization. 
The results demonstrate that \textit{Fastformer} is much more efficient than many Transformer models and can achieve quite competitive results in long text modeling.

The contributions of this paper are summarized as follows:

\begin{itemize}
    \item We propose an additive attention based  Transformer named \textit{Fastformer}.
    To the best of our knowledge, \textit{Fastformer} is the most efficient Transformer architecture.
    \item We propose to model the interaction between global contexts and  token representations via element-wise product, which can help fully model context information in an efficient way. 
    \item Extensive experiments on five datasets show that \textit{Fastformer} is much more efficient than many Transformer models and can achieve competitive performance.
\end{itemize}
\section{Related Work}

\subsection{Transformer and Self-Attention}

The Transformer model is built upon multi-head self-attention, which can effectively model the contexts within a sequence by capturing the interactions between the inputs at each pair of positions~\cite{vaswani2017attention}.
An $h$-head self-attention mechanism can be formulated as follows:
\begin{equation}
\begin{aligned}
    &\rm{MultiHead}(\mathbf{Q}, \mathbf{K}, \mathbf{V}) =\\ 
    &\rm{Concat}(\rm{head}_1, \rm{head}_2, ..., \rm{head}_h)\mathbf{W}^O,
\end{aligned}
\end{equation}
where $\mathbf{Q}$, $\mathbf{K}$, $\mathbf{V}\in \mathbb{R}^{N\times d}$ are the input query, key and value matrices, $N$ is the sequence length, $d$ is the hidden dimension in each attention head, and $\mathbf{W}^O\in \mathbb{R}^{hd\times d}$ is a linear transformation parameter matrix.
The representation learned by each attention head is formulated as follows:
\begin{equation}
\begin{aligned}
    \rm{head}_i&= \rm{Attention}(\mathbf{Q}\mathbf{W}^Q_i,\mathbf{K}\mathbf{W}^K_i,\mathbf{V}\mathbf{W}^V_i)\\ 
    &=\rm{softmax}(\frac{\mathbf{Q}\mathbf{W}^Q_i(\mathbf{K}\mathbf{W}^K_i)^T}{\sqrt{d}})\mathbf{V}\mathbf{W}^V_i,
\end{aligned}
\end{equation}
where $\mathbf{W}^Q_i$, $\mathbf{W}^K_i$, $\mathbf{W}^V_i\in \mathbb{R}^{d\times d}$ are learnable parameters.
From this formula, we can see that the computational complexity is quadratic to the sequence length $N$.
It has become a bottleneck for Transformers to handle long sequences.

\subsection{Efficient Transformer}

In recent years, there are many approaches to improving the efficiency of Transformer architecture~\cite{tay2020efficient}.
Some methods use sparse attention mechanisms to reduce the complexity of self-attention~\cite{child2019generating,beltagy2020longformer,zaheer2020big,zhang2021pooling}.
For example, Longformer~\cite{beltagy2020longformer} uses sliding window attention to attend local contexts and uses global attention on a few pre-selected input locations to capture global contexts.
BigBird~\cite{zaheer2020big} combines local attention, global attention at certain positions and random attention on several randomly selected token pairs.
However, these methods may need a relative larger  number of attended tokens to reduce the performance degradation on longer sequences, which usually leads to a limited speed-up ratio.

Another way is using hashing technique to accelerate self-attention computation.
For example, Reformer~\cite{Kitaev2020reformer} uses a multi-round hashing scheme to put similar representations into same buckets when computing self-attention, which can theoretically reduce the complexity to $O(N\log(N))$.
However, the computational complexity constant of Reformer is quite large, making it inefficient in processing common sequences with rather limited lengths.

There are also several methods that aim to reduce the computational cost by computing approximated  self-attention~\cite{choromanski2020masked,wang2020linformer,tay2021synthesizer}.
For instance, Linformer~\cite{wang2020linformer} projects the attention key and value into low-rank matrices to approximate the self-attention mechanism.
Linear Transformer~\cite{katharopoulos2020transformers} uses a kernel-based formulation of self-attention and the associative
property of matrix multiplication to approximate the dot-product attention.
However, these methods approximate self-attention in a context-agnostic manner, which may not be optimal for text modeling. 
In addition, they still bring heavy computational cost when the sequence length is very long.
Different from the aforementioned methods, \textit{Fastformer} uses additive attention to model global contexts and uses element-wise product to model the interaction between each input representation and global contexts, which can greatly reduce the computational cost and meanwhile effectively capture contextual information.

\begin{figure*}[!t]
  \centering 
      \includegraphics[width=0.64\linewidth]{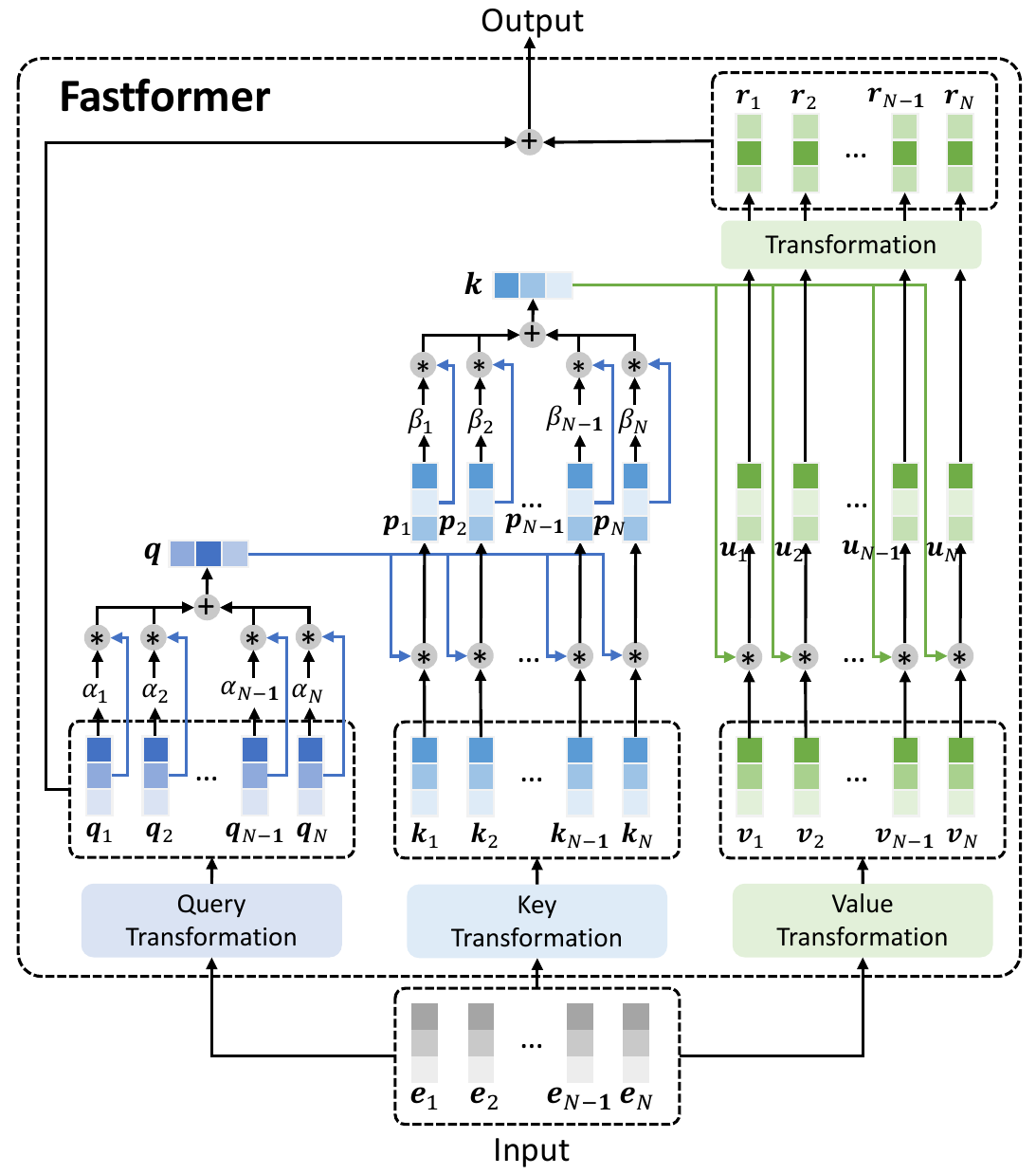}
  \caption{The architecture of \textit{Fastformer}.}\label{fig.model}

\end{figure*}

\section{Fastformer}\label{sec:Model}

In this section, we introduce our \textit{Fastformer} approach based on additive attention.
The architecture of \textit{Fastformer} is shown in Fig.~\ref{fig.model}.
It first uses additive attention mechanism to summarize the query sequence into a global query vector, next models the interaction between the global query vector and attention keys with element-wise product and summarize keys into a global key vector via additive attention, then models the interactions between global key and attention values via element-wise product and uses a linear transformation to learn global context-aware attention values, and finally adds them with the attention query to form the final output.
In this way, the computational complexity can be reduced to linearity, and the contextual information in the input sequence can be effectively captured.
Next, we introduce the details of \textit{Fastformer} in the following section.

\subsection{Architecture}

The \textit{Fastformer} model first transforms the input embedding matrix into the query, key and value sequences.
The input matrix is denoted as $\mathbf{E}\in \mathbb{R}^{N\times d}$, where $N$ is the sequence length and $d$ is the hidden dimension.
Its subordinate vectors are denoted as $[\mathbf{e}_1, \mathbf{e}_2, ..., \mathbf{e}_N]$.
Following the standard Transformer, in each attention head\footnote{Different attention heads use the same formulation but different parameters.} we use three independent linear transformation layer to transform the input into the attention query, key and value matrices $\mathbf{Q}, \mathbf{K}, \mathbf{V}\in \mathbb{R}^{N\times d}$, which are written as $\mathbf{Q}=[\mathbf{q}_1, \mathbf{q}_2, ..., \mathbf{q}_N]$, $\mathbf{K}=[\mathbf{k}_1, \mathbf{k}_2, ..., \mathbf{k}_N]$ and $\mathbf{V}=[\mathbf{v}_1, \mathbf{v}_2, ..., \mathbf{v}_N]$, respectively.

Next, modeling the contextual information of the input sequence based on the interactions among attention query, key and value is a critical problem for Transformer-like architectures.
In the vanilla Transformer, dot-product attention mechanism is used to fully model the interaction between query and key.
Unfortunately, its quadratic complexity makes it inefficient in long sequence modeling.
A potential way to reduce the computational complexity is to summarize the attention matrices (e.g., query) before modeling their interactions.
Additive attention is a form of attention mechanism that can efficiently summarize important information within a sequence in linear complexity.
Thus, we first use  additive attention to summarize the query matrix into a global query vector $\mathbf{q}\in \mathbb{R}^{d}$, which condenses the global contextual information in the attention query.
More specifically, the attention weight $\alpha_i$ of the $i$-th query vector is computed as follows:
\begin{equation}
\alpha_i=\frac{\exp(\mathbf{w}_q^T\mathbf{q}_i/\sqrt{d})}{\sum_{j=1}^N\exp(\mathbf{w}_q^T\mathbf{q}_j/\sqrt{d})},
\end{equation}
where $\mathbf{w}_q\in \mathbb{R}^{d}$ is a learnable parameter vector.
The global attention query vector is computed as follows:
\begin{equation}
\mathbf{q}=\sum_{i=1}^N\alpha_i\mathbf{q}_i.
\end{equation}

Then, a core problem in \textit{Fastformer} is how to model the interaction between the summarized global query vector and the key matrix.
There are several intuitive options, such as adding or concatenating the global query to each vector in the key matrix.
However, they cannot differ the influence of the global query on different keys, which is not beneficial for context understanding.
Element-wise product is an effective operation to model the non-linear relations between two vectors~\cite{wang2017item}.
Thus, we use the element-wise product between the global query vector and each key vector to model their interactions and combine them into a global context-aware key matrix.
We denote the $i$-th vector in this matrix as $\mathbf{p}_i$, which is formulated as $\mathbf{p}_i=\mathbf{q} * \mathbf{k}_i$ (the symbol $*$ means element-wise product).
In a similar way, we use additive attention mechanism to summarize the global context-aware key matrix due to efficiency reasons.
The additive attention weight of its $i$-th vector is computed as follows:
\begin{equation}
\beta_i=\frac{\exp(\mathbf{w}_k^T\mathbf{p}_i/\sqrt{d})}{\sum_{j=1}^N\exp(\mathbf{w}_k^T\mathbf{p}_j/\sqrt{d})},
\end{equation}
where $\mathbf{w}_k\in \mathbb{R}^{d}$ is the attention parameter vector.
The global key vector $\mathbf{k}\in \mathbb{R}^{d}$ is further computed as follows:
\begin{equation}
\mathbf{k}=\sum_{i=1}^N\beta_i\mathbf{p}_i.
\end{equation}

Finally, we model the interaction between attention value matrix and the global key vector for better context modeling.
Similar with the query-key interaction modeling, we also perform element-wise product between the global key and each value vector to compute a key-value interaction vector $\mathbf{u}_i$, which is formulated as $\mathbf{u}_i=\mathbf{k}*\mathbf{v}_i$.
Motivated by the vanilla Transformer, we apply a linear transformation layer to each key-value interaction vector to learn its hidden representation.
The output matrix from this layer is denoted as $\mathbf{R}=[\mathbf{r}_1, \mathbf{r}_2, ..., \mathbf{r}_N]\in \mathbb{R}^{N\times d}$.
This matrix is further added together with the query matrix to form the final output of \textit{Fastformer}.
Note that the output matrix from each attention head is concatenated along the hidden dimension axis.

We can see that in \textit{Fastformer}, each key and value vector can interact with the global query or key vector to learn contextualized representations.
By stacking multiple \textit{Fastformer} layers, we can fully model contextual information.
Motivated by the weight sharing techniques used in~\cite{wang2020linformer}, we share the value and query transformation parameters to reduce the memory cost.
In addition, we share the parameters across different  \textit{Fastformer} layers to further reduce the parameter size and mitigate the risk of overfitting.

\subsection{Complexity Analysis}

In this section, we analyze the computational complexity of \textit{Fastformer}.
For the additive attention networks to learn global query and key vectors, their time and memory cost are both $O(N\cdot d)$, and their total number of additional parameters is $2hd$ ($h$ is the attention head number).
In addition, the time cost and memory cost of element-wise product is also $O(N\cdot d)$, the total complexity is $O(N\cdot d)$, which is much more efficient than the standard Transformer with $O(N^2 \cdot d)$ complexity.\footnote{Note that the complexity of linear transformations is not taken into account by many prior works, and we also do not consider their effects on computational costs.}
If the weight sharing technique is used, the total parameter size of \textit{Fastformer} per layer is $3hd^2+2hd$.
Compared with Transformer with at least $4hd^2$ parameters\footnote{Including the query, key, value and output transformation matrices. The parameters in the bias term, feed-forward network and layer normalization are not counted.}, \textit{Fastformer} also uses fewer parameters.
These analysis results demonstrate the theoretical efficiency of \textit{Fastformer}.
\section{Experiments}\label{sec:Experiments}

\begin{table}[!t]
\centering
\resizebox{1\linewidth}{!}{ 
\begin{tabular}{ccccccc}
 \Xhline{1.5pt}
\multicolumn{1}{c}{\textbf{Dataset}} & \multicolumn{1}{c}{\textbf{\#Train}} & \multicolumn{1}{c}{\textbf{\#Val}} & \multicolumn{1}{c}{\textbf{\#Test}} & \multicolumn{1}{c}{\textbf{Avg. len.}} & \multicolumn{1}{l}{\textbf{\#Class}} \\ \hline

Amazon                               & 40.0k                                & 5.0k                               & 5.0k                                & 133.4                                                            & 5                                    \\
IMDB                                 & 108.5k                               & 13.6k                              & 13.6k                               & 385.7                                                                     & 10                                   \\
MIND                                 & 128.8k                               & 16.1k                              & 16.1k                               & 505.4                                                               & 18                                   \\  \Xhline{1.5pt}
\end{tabular}
}
\caption{Statistics of sentiment and news topic classification datasets.}\label{dataset}
\end{table}

\begin{table}[!t]
\centering
\resizebox{1\linewidth}{!}{ 
\begin{tabular}{lclc}
\Xhline{1.5pt}
\#News  &  161,013  & \#Users    & 1,000,000  \\
\#Train impression  &  2,232,748   &   \#Val impression  &  376,471  \\
\#Test impression  &  2,370,727    & Avg. click his. len. & 37.1\\
\Xhline{1.5pt}
\end{tabular}
}
\caption{Statistics of MIND dataset for the news recommendation task.}\label{dataset2}
\end{table}

\begin{table}[!t]
\centering
\resizebox{1\linewidth}{!}{ 
\begin{tabular}{cccccc}
 \Xhline{1.5pt}
\multicolumn{1}{c}{\textbf{Dataset}} & \multicolumn{1}{c}{\textbf{\#Train}} & \multicolumn{1}{c}{\textbf{\#Val}} & \multicolumn{1}{c}{\textbf{\#Test}} & \multicolumn{1}{c}{\textbf{Avg. doc/summary len.}} \\ \hline

CNN/DailyMail                               & 287.1k                                & 13.4k                               & 11.5k                                & 781/56                                    \\
PubMed                                 & 108.5k                               & 13.6k                              & 13.6k                               & 3016/203                                   \\ \Xhline{1.5pt}
\end{tabular}
}
\caption{Statistics of the text summarization datasets.}\label{dataset3}
\end{table}
\begin{table*}[!t]
\resizebox{0.98\textwidth}{!}{
\begin{tabular}{lcccccc}
\Xhline{1.5pt}
\multicolumn{1}{c}{\multirow{2}{*}{Methods}} & \multicolumn{2}{c}{Amazon}      & \multicolumn{2}{c}{IMDB} & \multicolumn{2}{c}{MIND} \\ \cline{2-7} 
\multicolumn{1}{c}{}                         & Accuracy       & Macro-F        & Accuracy    & Macro-F    & Accuracy    & Macro-F    \\ \hline
Transformer                                  & 65.32$\pm$0.35 & 42.31$\pm$0.33 & 52.04$\pm$0.50       & 42.69$\pm$0.47     & 80.90$\pm$0.20       & 60.02$\pm$0.21      \\
Longformer                                   & 65.45$\pm$0.39 & 42.48$\pm$0.44 & 52.21$\pm$0.36       & 43.36$\pm$0.38     & 81.36$\pm$0.21       & 62.59$\pm$0.23      \\
BigBird                                     & 66.14$\pm$0.42 & 42.96$\pm$0.40 & 53.23$\pm$0.46       & 44.03$\pm$0.44     & 81.93$\pm$0.24       & 63.58$\pm$0.26      \\ 
Linformer                                     & \textbf{66.20}$\pm$0.49 & 43.13$\pm$0.48 & 53.17$\pm$0.59       & 44.34$\pm$0.57     & 82.16$\pm$0.28       & 63.77$\pm$0.30      \\ 
Linear Transformers                                     & 66.12$\pm$0.42 & 43.04$\pm$0.44 & 53.09$\pm$0.47       & 44.30$\pm$0.49     & 82.25$\pm$0.23       & 63.81$\pm$0.22      \\ 
Poolingformer                                     & 66.05$\pm$0.44 & 43.00$\pm$0.45 & 53.78$\pm$0.51       & 44.52$\pm$0.50     & \textbf{82.46}$\pm$0.24       & \textbf{64.10}$\pm$0.26      \\ 
\hline
Fastformer                               & 66.13$\pm$0.29 & \textbf{43.23}$\pm$0.30 & \textbf{54.10}$\pm$0.42       & \textbf{44.65}$\pm$0.44     & 82.34$\pm$0.19       & 63.89$\pm$0.20      \\  \Xhline{1.5pt}
\end{tabular}
}
\caption{The results of different methods in the sentiment and topic classification tasks. Best average scores are highlighted.} \label{table.performance} 
\end{table*}

\subsection{Datasets and Experimental Settings}

We conduct extensive experiments on five benchmark datasets for different tasks.
Their details are introduced as follows.
The first one is Amazon~\cite{he2016ups} (we use the Electronics domain)\footnote{https://jmcauley.ucsd.edu/data/amazon/}, which is a widely used dataset for review rating prediction.
The second one is IMDB~\cite{diao2014jointly}.\footnote{https://github.com/nihalb/JMARS}
It is a benchmark dataset for movie review rating prediction.
The third one is MIND~\cite{wu2020mind}\footnote{https://msnews.github.io/}, which is a large-scale English dataset for news recommendation and intelligence.
We perform two tasks on this dataset, i.e., the news topic classification task based on news body and personalized news recommendation task based on the relevance between candidate news and user interests inferred from historical clicked news.
The fourth one is CNN/DailyMail dataset~\cite{hermann2015teaching} (denoted as CNN/DM), which is a widely used benchmark dataset for text summarization. 
The fifth one is PubMed~\cite{cohan2018discourse}, which is another benchmark text summarization dataset with much longer documents lengths.
The detailed statistical information of the datasets introduced above are shown in Tables~\ref{dataset},~\ref{dataset2} and~\ref{dataset3}.

In our experiments, we use Glove~\cite{pennington2014glove} embeddings to initialize token embedding matrix.
%In \textit{Fastformer} and other baseline methods we use two layers.\footnote{We explored using more Transformer layers for baselines while no significant improvement is observed.}
To obtain the embeddings in the classification and news recommendation tasks, we apply an additive attention network to convert the matrix output by \textit{Fastformer} into an embedding.
In addition, in the news recommendation task, following~\cite{wu2019nrms} we use \textit{Fastformer} in a hierarchical way to first learn news embeddings from news titles and then learn user embeddings from the embeddings of historical clicked news.
%The total hidden dimension is 256, which means that the number of attention head $h$ is 8 and the hidden dimension  of each head $d$ is 32.\footnote{We found that using larger hidden dimension did not bring notable performance improvement.}
We use Adam~\cite{kingma2014adam} for model optimization.
More detailed experimental settings are included in Appendix.
We run our experiments on an Nvidia Tesla V100 GPU with 32GB memory.
We repeat each experiment 5 times and report the average performance as well as the standard deviations.
On the classification tasks, we use accuracy and macro-F scores as performance metrics.
On the news recommendation task, following~\cite{wu2020mind} we use AUC, MRR, nDCG@5 and nDCG@10 as the metrics.
On the text summarization tasks, we use the ROUGE-1, ROUGE-2 and ROUGE-L metrics (denoted as R-1, R-2 and R-L) to evaluate the generated summaries.

\subsection{Effectiveness Comparison}

First, we compare the performance of \textit{Fastformer} with many baseline methods, including:
(1) \textit{Transformer}~\cite{vaswani2017attention}, the vanilla Transformer;
(2) \textit{Longformer}~\cite{beltagy2020longformer}, a   Transformer variant with sparse attention. It combines sliding window attention and global attention to model local and global contexts;
(3) \textit{BigBird}~\cite{zaheer2020big}, an extension of  \textit{Longformer}, which incorporates sparse random attention mechanism;
(4) \textit{Linformer}~\cite{wang2020linformer}, a Transformer variant with linear complexity, which use low-dimensional key and value matrices to compute approximated self-attention; 
(5) \textit{Linear Transformer}~\cite{katharopoulos2020transformers}, another linear complexity Transformer using kernel functions to approximate self-attention mechanism;
(6) \textit{Poolingformer}~\cite{zhang2021pooling}, a hierarchical architecture that first uses sliding window self-attention to capture short-range contexts and then uses pooling self-attention to capture long-range contexts.

The performance of these methods on the three classification datasets are compared in Table~\ref{table.performance}.
From the results, we find that efficient Transformer variants usually outperform the standard Transformer model.
This is because the quadratic computational cost of vanilla Transformer limits the maximum sequence length can be handled, and many useful contexts are lost when truncating the input text sequence.
\textit{Fastformer} can achieve competitive or better performance than other efficient Transformer variants in both long and short text modeling.
This is because \textit{Fastformer} can effectively model global contexts and their relationship to different tokens, which can help understand context information accurately.

\begin{table}[!t]
 
 \resizebox{1.0\linewidth}{!}{
\begin{tabular}{lcccc}
\Xhline{1.5pt}
\textbf{Methods} & \textbf{AUC}   & \textbf{MRR}   & \multicolumn{1}{l}{\small{\textbf{nDCG@5}}} & \multicolumn{1}{l}{\small{\textbf{nDCG@10}}} \\ \hline
NRMS             & 68.18          & 33.29          & 36.31             & 42.20          \\
FIM             & 68.31          & 33.42          & 36.47             & 42.35          \\
PLM-NR  & 70.64 & 35.39 & 38.71    & 44.38 \\ \hline
Transformer       & 68.22          & 33.32          & 36.35             & 42.23 \\ 
Longformer       & 67.98          & 33.04          & 36.18             & 42.06 \\ 
BigBird      & 68.14          & 33.28          & 36.30             & 42.18 \\ 
Linformer       & 68.02          & 33.19          & 36.22             & 42.10 \\ 
Linear Transformers       & 67.76          & 32.94          & 36.16             & 41.97 \\
Poolingformer       & 68.54          & 33.60          & 36.69             & 42.60 \\  \hline
Fastformer    & 69.11          & 34.25          & 37.26             & 43.38 \\

Fastformer+PLM-NR  & 71.04          & 35.91          & 39.16             & 45.03 \\
Fastformer+PLM-NR*  & 72.68          & 37.45          & 41.51             & 46.84 \\
\Xhline{1.5pt}
\end{tabular}
}
\caption{Performance of different methods in the news recommendation task. *Ensemble of five independently trained models, which is the 1$^{st}$ ranked result on the MIND leaderboard.} \label{table.performance2} 
\end{table}

We also compare the performance of different methods in the news recommendation task. 
We add three recent news recommendation methods to the comparison, including: (1) \textit{NRMS}~\cite{wu2019nrms}, which uses multi-head self-attention networks to learn news and user representations; (2) \textit{FIM}~\cite{wang2020fine}, a fine-grained interest matching method for personalized news recommendation; (3) \textit{PLM-NR}~\cite{wu2021plm}, empowering news recommendation with pre-trained language models. In \textit{PLM-NR}, we use the best performed UniLM~\cite{bao2020unilmv2} empowered model.
In addition, we explore to replace the user encoder in \textit{PLM-NR} with \textit{Fastformer}.
The results are shown in Table~\ref{table.performance2}.
We can see that among different Transformer architectures, \textit{Fastformer} achieves the best performance, and it also outperforms its basic \textit{NRMS} model.
In addition, \textit{Fastformer} can further improve the performance of \textit{PLM-NR}, and the ensemble model achieves the best results on the MIND leaderboard\footnote{https://msnews.github.io/}.
These results show that \textit{Fastformer} is not only effective in text modeling, but also effective in understanding user interest.

\begin{table}[!t]
\resizebox{0.98\linewidth}{!}{
\begin{tabular}{lcccccc}
\Xhline{1.5pt}
\multicolumn{1}{c}{\multirow{2}{*}{Method}} & \multicolumn{3}{c}{CNN/DailyMail} & \multicolumn{3}{c}{PubMed} \\ \cline{2-7} 
\multicolumn{1}{c}{}                        & R-1       & R-2       & R-L       & R-1     & R-2     & R-L    \\ \hline
Transformer                                 &   38.52        &  16.04         &   35.87        &   34.26      &  11.88       &  31.64      \\
Longformer                                  &   37.89        &  15.46         &   35.19        &   36.92      &  14.34       &  33.75      \\
BigBird                                     &   38.31        &  15.78         &   35.60        &   37.73      &  14.99       &  34.51      \\
Linformer                                   &   37.96        &  15.58         &   35.34        &   37.22      &  14.48       &  34.02      \\
Linear Transformer                          &   37.24        &  14.87         &   34.64        &   36.43      &  13.80       &  33.21      \\
Poolingformer                               &   \textbf{38.58}        &  16.16         &   36.17        &   37.82      &  15.15       &  34.63      \\ \hline
Fastformer                                  &   38.54        &  \textbf{16.22}         &   \textbf{36.21}        &   \textbf{38.09}      &  \textbf{15.44}       &  \textbf{34.81}      \\ \Xhline{1.5pt}
\end{tabular}
}
\caption{Performance of different methods in the text summarization task. Best scores are highlighted.} \label{table.performance3} 
\end{table}

We further conduct experiments on the text summarization tasks to verify the effectiveness of \textit{Fastformer} in natural language generation.
The results are shown in Table~\ref{table.performance3}.
We find that on the CNN/DM dataset, many efficient Transformer variants (except \textit{Poolingformer} and \textit{Fastformer}) are inferior to the vanilla Transformer.
This is because sparse attention based method such as \textit{Longformer} and \textit{BigBird} cannot fully model the document contexts, and approximated self-attention based methods such as \textit{Linformer} and \textit{Linear Transformer} cannot effectively consider context information in the approximation.
Since the summaries in the CNN/DM dataset have short lengths, they may  be less effective than vanilla Transformer when the same sequence length is used.
\textit{Fastformer} can achieve the best performance in most metrics, which shows the advantage of \textit{Fastformer} in natural language generation.

\begin{table}[t]
\centering
%\resizebox{1.0\linewidth}{!}{
\begin{tabular}{lc}
\Xhline{1.5pt}
\multicolumn{1}{c}{\textbf{Method}} & \multicolumn{1}{c}{\textbf{Complexity}} \\ \hline
Transformer                         & $O(N^2\cdot d)$                          \\
Longformer                          & $O(N\cdot K \cdot d)$                                             \\
BigBird                             & $O(N\cdot K \cdot d)$                                             \\
Linformer                             & $O(N \cdot d/\epsilon^2)$                                             \\
Linear Transformer                             & $O(N  \cdot d^2)$                                             \\
Poolingformer                             & $O(N\cdot d \cdot w)$                                             \\
Fastformer                      & $O(N\cdot d)$                         \\ \Xhline{1.5pt}
\end{tabular}
%}
\caption{Asymptotic computational complexity of different methods. $N$ is the sequence length, $K$ is the average number of tokens to be attended by per token, $d$ is the hidden dimension, $\epsilon$ is the attention matrix approximation error in Linformer, and $w$ is the window size in Poolingformer.}\label{complexity}
\end{table}

\begin{figure*}[!t]
  \centering
   \subfigure[Inference.]{
      \includegraphics[width=0.48\linewidth]{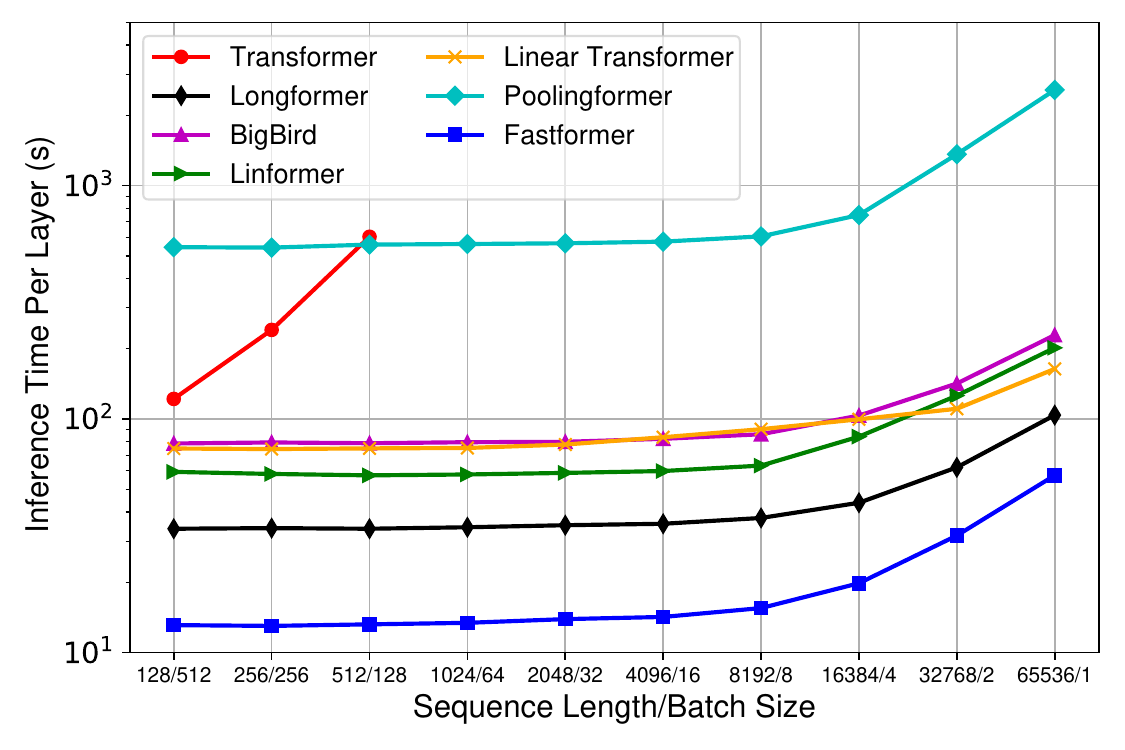}
  \label{fig.len2}
  } 
   \subfigure[Training.]{
      \includegraphics[width=0.48\linewidth]{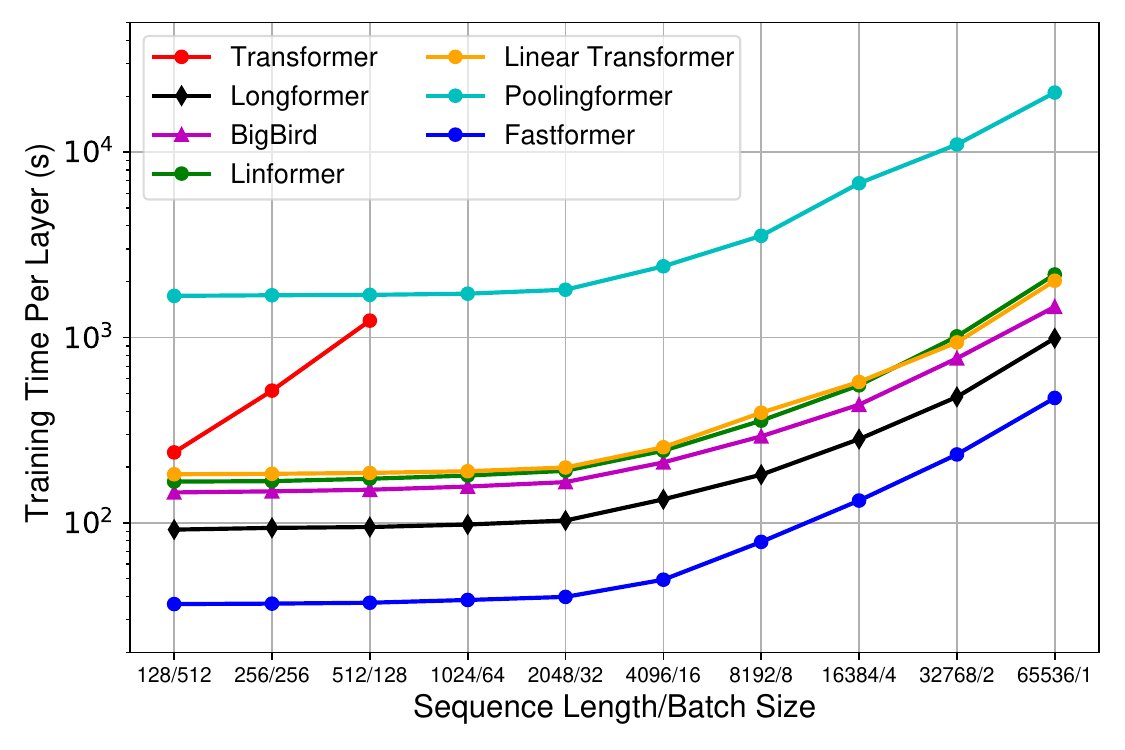}
  \label{fig.len3}
  }
  \caption{Training and inference speed of different methods. The y-axis (time) is in logarithmic scale.}\label{fig.speed}
\end{figure*}

\subsection{Efficiency Comparison}

In this section, we evaluate the efficiency of different methods.
We first compare the theoretical computational complexity of these methods in Table~\ref{complexity}.\footnote{We exclude the complexity of the linear Transformation applied after the input and before the output.}
The complexity of vanilla Transformer is $O(N^2\cdot d)$, while other compared methods have linear complexity with respect to the input sequence length.
However, the complexity of \textit{Longformer} and \textit{BigBird} depends on the average number of tokens to be attended by each token, \textit{Linformer} has a complexity depending on the square of the approximation error, the complexity of \textit{Linear Transformer} has a quadratic term of hidden dimension, and the complexity of \textit{Poolingformer} depends on its window size.
Different from them, the complexity of \textit{Fastformer} only depends on the sequence length and the hidden dimension, and it has the least complexity among compared methods.
This result shows that \textit{Fastformer} is efficient in theory.

Then, we conduct experiments to measure the real training and inference cost of different methods.\footnote{We use the model configurations in the news topic classification task.}
Motivated by prior works~\cite{katharopoulos2020transformers,wang2020linformer}, we vary the sequence length from 128 to 65535 and scale the batch size inversely with the sequence length.
We generate pseudo samples with random tokens and fix the token embeddings to better measure the computational cost of different methods.
The results are  shown in Fig.~\ref{fig.speed}.\footnote{The maximum sequence length of Transformer is limited by the GPU memory.}
We find that Transformer is inefficient when the sequence length is relatively long (e.g., 512).
In addition, we find that although \textit{Poolingformer} has linear complexity in theory, it is inefficient in practice.
This is because it uses a large window size (e.g., 256) to compute pooling weight in a convolution-like way, which leads to a very large constant term of the computational cost.
Besides, \textit{Fastformer} is much more efficient than other linear complexity Transformer variants in terms of both training and inference time.
These results verify the efficiency of \textit{Fastformer}.

\subsection{Influence of Interaction Function}

\begin{figure}[!t]
  \centering
  \subfigure[Amazon.]{
    \includegraphics[height=1.6in]{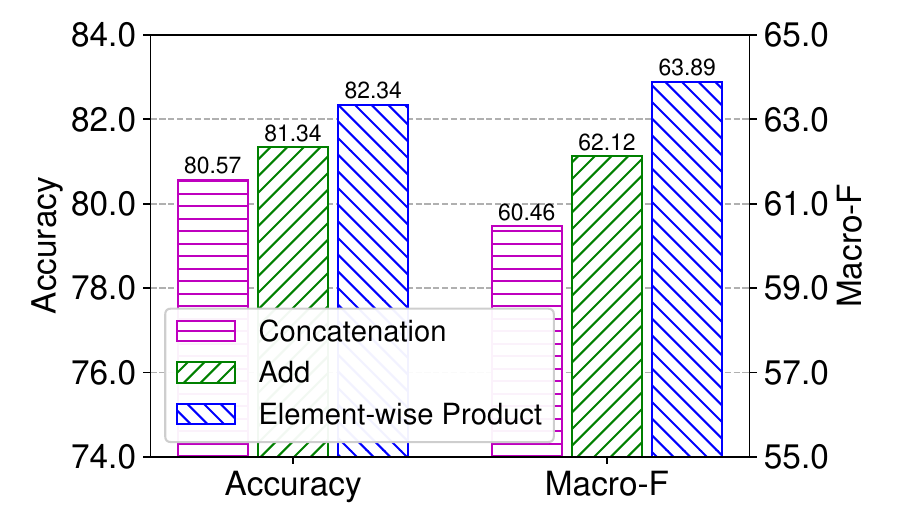}
  \label{fig.ab1}
  }
   \subfigure[IMDB.]{
      \includegraphics[height=1.6in]{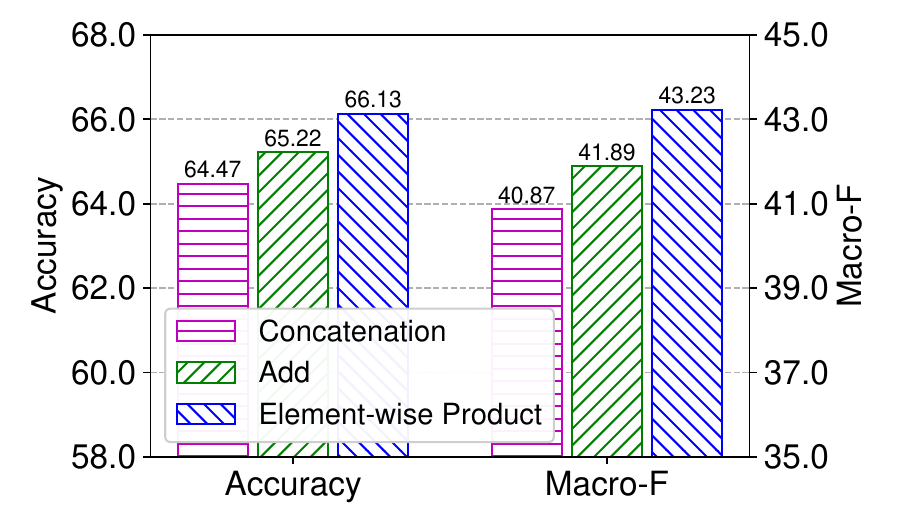}
  \label{fig.ab2}
  }
   \subfigure[MIND.]{
      \includegraphics[height=1.6in]{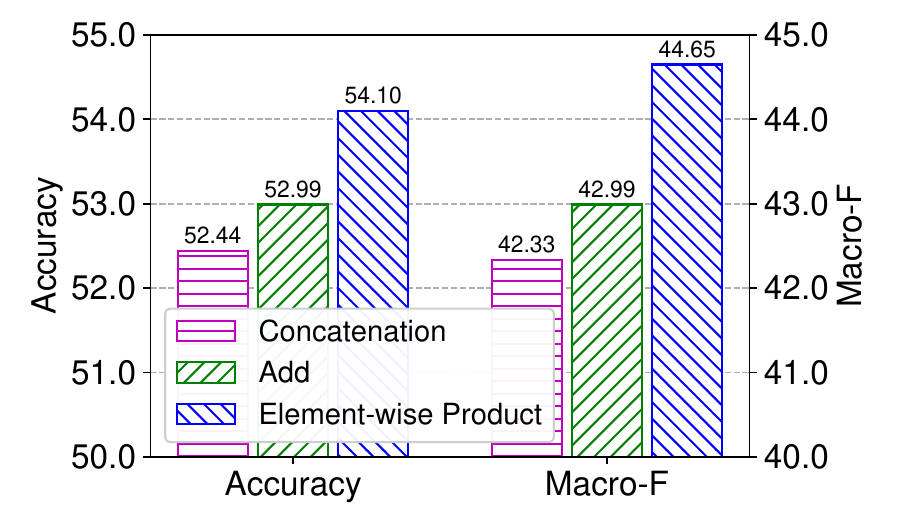}
  \label{fig.ab3}
  }
  \caption{Influence of different combination functions.}\label{fig.ab}

\end{figure}

Next, we study the influence of using different functions to model the interactions among query, key and value in \textit{Fastformer}.
We compare the performance of \textit{Fastformer} and its variants using the add or concatenation functions to combine the global query/key vector with the vectors in the key/value matrix.
The results are shown in Fig.~\ref{fig.ab}.
We find concatenating is not a good option for \textit{Fastformer}.
This is because simply concatenating two vectors cannot consider the interactions between them.
In addition, we find adding is not optimal.
This is because the add function can only model the linear interactions between two vectors, which may also be insufficient to learn accurate context representations.
Different from concatenation and add, element-wise product can model the non-linear interactions between two variables, which may help model the complex contexts in long sequences.

\begin{figure}[!t]
  \centering
  \subfigure[Amazon.]{
    \includegraphics[height=1.6in]{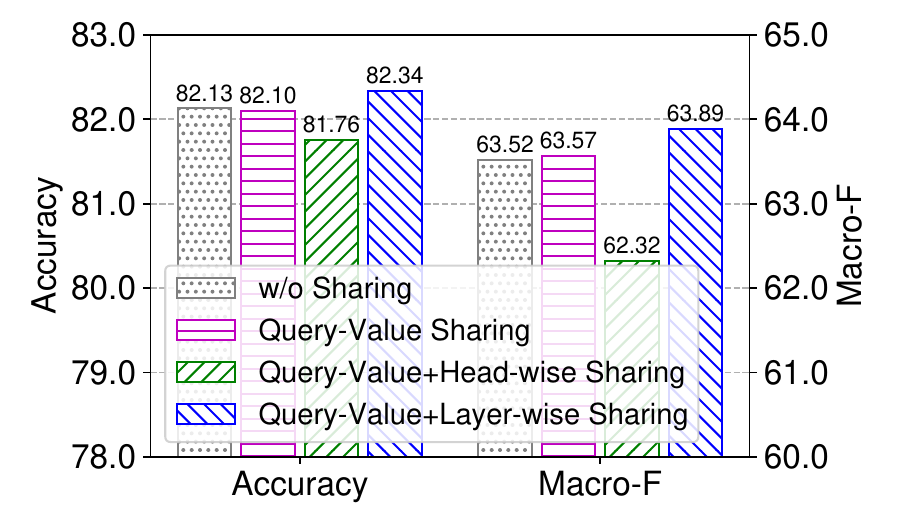}
  \label{fig.eff1}
  }
   \subfigure[IMDB.]{
      \includegraphics[height=1.6in]{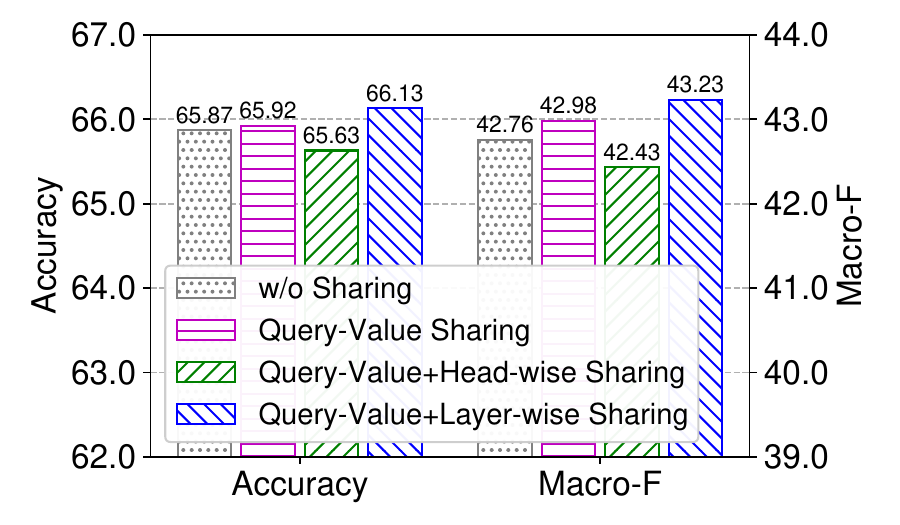}
  \label{fig.eff2}
  }
   \subfigure[MIND.]{
      \includegraphics[height=1.6in]{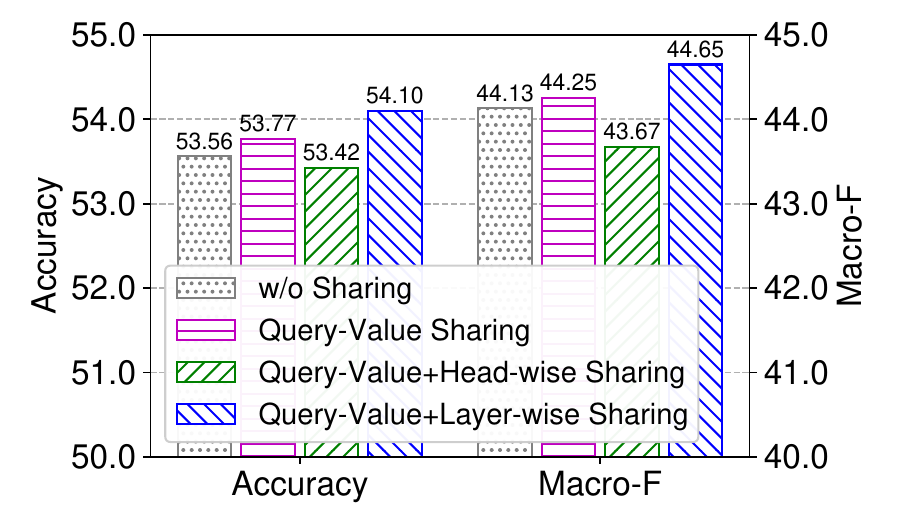}
  \label{fig.eff3}
  }
  \caption{Influence of different parameter sharing strategies.}\label{fig.eff}

\end{figure}

\subsection{Influence of Parameter Sharing}

Then, we study the influence of different parameter sharing techniques on the performance of \textit{Fastformer}, including sharing query and value transformation matrices, sharing the parameters across different attention heads, and sharing parameters across different layers.\footnote{We do not observe significant differences in terms of time cost when using different parameter sharing methods, and we only compare the performance here.}
The results are shown in Fig.~\ref{fig.eff}.
We find that using query-value parameter sharing can achieve similar or slightly better performance than the \textit{Fastformer} model without any parameter sharing techniques.
Thus, it is favorable to reduce the parameter size by sharing the query and value transformation matrix.
In addition, we find head-wise parameter sharing will lead to  notable performance drops.
This is because different attention heads are expected to capture different patterns of contexts, and sharing their parameters is not beneficial for context modeling.
Moreover, we find incorporating layer-wise sharing method can further improve the model performance.
This is because parameter sharing among different layers can mitigate the risk of overfitting~\cite{Lan2020albert}.
Thus, in \textit{Fastformer} we incorporate both query-value sharing and layer-wise sharing strategies to improve the model performance and meanwhile reduce the model parameter size.

\subsection{Applications of Fastformer}

We further apply \textit{Fastformer} to a downstream Ad CVR prediction task.
We conduct experiments on a large-scale Ad CVR prediction data collected from Bing Ads, which contains user behaviors such as search query,  webpage browsing and the conversion of clicked Ads.
The task is a binary classification task by predicting whether a clicked Ad leads to a conversion.
The dataset contains 52.2m training samples, 387k validation samples and 611k test samples (logs in the last week are for test, and the rest are for training and validation).
Similar to news recommendation, we first use a \textit{Transformer} or \textit{Fastformer} to learn ad/behavior embedding, and then use another \textit{Transformer} or \textit{Fastformer} to learn user embedding from behavior embeddings.
The prediction AUC, training memory utilization and local inference latency of \textit{Transformer} and \textit{Fastformer} based methods are shown in Table~\ref{cvr}.
The results show the effectiveness and efficiency of \textit{Fastformer} in Ad CVR prediction.\footnote{The latency improvement scale is relatively smaller than in Fig.~\ref{fig.speed} because the sequences are much shorter and there are multiple additional MLPs in the model.}

\begin{table}[h]
\centering
\resizebox{1.0\linewidth}{!}{
\begin{tabular}{llll}
\hline
            & AUC   & Memory        & Latency         \\ \hline
Transformer & 0.7299         & 30GB          & 176ms         \\
Fastformer  & 0.7394(+1.3\%) & 25GB(-16.7\%) & 163ms(-7.2\%) \\ \hline
\end{tabular}
}
  \caption{Accuracy, memory cost and inference speed comparison.}\label{cvr}

\end{table}

\section{Conclusion and Future Work}\label{sec:Conclusion}

In this paper, we propose \textit{Fastformer}, which is a  Transformer variant based on additive attention that can handle long sequences efficiently with linear complexity.
In \textit{Fastformer}, we first use additive attention to summarize the query matrix into a global query vector.
Next, we combine it with each key vector via element-wise product to learn global context-aware key matrix, and we further summarize it into a global key vector via additive attention.
We then model the interactions between global context-aware key and the value to learn global context-aware attention value, which is further combined with the query to form the final output.
Extensive experiments on five benchmark datasets show that \textit{Fastformer} is much more efficient than many existing Transformer models and meanwhile can achieve competitive or even better performance in long text modeling.

In our future work, we plan to pre-train \textit{Fastformer}-based language models to better empower NLP tasks with long document modeling.
In addition, we will explore applying \textit{Fastformer} to other scenarios such as e-commerce recommendation and Ads CTR prediction to improve user modeling based on long user behavior sequences.

%\section*{Acknowledgments}
%This work was supported by the National Natural Science Foundation of China under Grant numbers U1936216, U1936208, U1836204, and U1705261.
%We are grateful to Xing Xie, Shaoyu Zhou, Dan Shen, and Zhisong Wang for their insightful comments and suggestions on this work.

\bibliographystyle{acl_natbib}
\bibliography{acl2021}

\begin{thebibliography}{29}
\expandafter\ifx\csname natexlab\endcsname\relax\def\natexlab#1{#1}\fi

\bibitem[{Bao et~al.(2020)Bao, Dong, Wei, Wang, Yang, Liu, Wang, Gao, Piao,
  Zhou et~al.}]{bao2020unilmv2}
Hangbo Bao, Li~Dong, Furu Wei, Wenhui Wang, Nan Yang, Xiaodong Liu, Yu~Wang,
  Jianfeng Gao, Songhao Piao, Ming Zhou, et~al. 2020.
\newblock Unilmv2: Pseudo-masked language models for unified language model
  pre-training.
\newblock In \emph{ICML}, pages 642--652. PMLR.

\bibitem[{Beltagy et~al.(2020)Beltagy, Peters, and
  Cohan}]{beltagy2020longformer}
Iz~Beltagy, Matthew~E Peters, and Arman Cohan. 2020.
\newblock Longformer: The long-document transformer.
\newblock \emph{arXiv preprint arXiv:2004.05150}.

\bibitem[{Bengio and LeCun(2015)}]{kingma2014adam}
Yoshua Bengio and Yann LeCun. 2015.
\newblock Adam: {A} method for stochastic optimization.
\newblock In \emph{ICLR}.

\bibitem[{Child et~al.(2019)Child, Gray, Radford, and
  Sutskever}]{child2019generating}
Rewon Child, Scott Gray, Alec Radford, and Ilya Sutskever. 2019.
\newblock Generating long sequences with sparse transformers.
\newblock \emph{arXiv preprint arXiv:1904.10509}.

\bibitem[{Choromanski et~al.(2020)Choromanski, Likhosherstov, Dohan, Song,
  Gane, Sarlos, Hawkins, Davis, Belanger, Colwell
  et~al.}]{choromanski2020masked}
Krzysztof Choromanski, Valerii Likhosherstov, David Dohan, Xingyou Song,
  Andreea Gane, Tamas Sarlos, Peter Hawkins, Jared Davis, David Belanger, Lucy
  Colwell, et~al. 2020.
\newblock Masked language modeling for proteins via linearly scalable
  long-context transformers.
\newblock \emph{arXiv preprint arXiv:2006.03555}.

\bibitem[{Cohan et~al.(2018)Cohan, Dernoncourt, Kim, Bui, Kim, Chang, and
  Goharian}]{cohan2018discourse}
Arman Cohan, Franck Dernoncourt, Doo~Soon Kim, Trung Bui, Seokhwan Kim, Walter
  Chang, and Nazli Goharian. 2018.
\newblock A discourse-aware attention model for abstractive summarization of
  long documents.
\newblock In \emph{NAACL-HLT}, pages 615--621.

\bibitem[{Devlin et~al.(2019)Devlin, Chang, Lee, and
  Toutanova}]{devlin2019bert}
Jacob Devlin, Ming-Wei Chang, Kenton Lee, and Kristina Toutanova. 2019.
\newblock Bert: Pre-training of deep bidirectional transformers for language
  understanding.
\newblock In \emph{NAACL-HLT}, pages 4171--4186.

\bibitem[{Diao et~al.(2014)Diao, Qiu, Wu, Smola, Jiang, and
  Wang}]{diao2014jointly}
Qiming Diao, Minghui Qiu, Chao-Yuan Wu, Alexander~J Smola, Jing Jiang, and
  Chong Wang. 2014.
\newblock Jointly modeling aspects, ratings and sentiments for movie
  recommendation (jmars).
\newblock In \emph{KDD}, pages 193--202.

\bibitem[{Dosovitskiy et~al.(2021)Dosovitskiy, Beyer, Kolesnikov, Weissenborn,
  Zhai, Unterthiner, Dehghani, Minderer, Heigold, Gelly
  et~al.}]{dosovitskiy2020image}
Alexey Dosovitskiy, Lucas Beyer, Alexander Kolesnikov, Dirk Weissenborn,
  Xiaohua Zhai, Thomas Unterthiner, Mostafa Dehghani, Matthias Minderer, Georg
  Heigold, Sylvain Gelly, et~al. 2021.
\newblock An image is worth 16x16 words: Transformers for image recognition at
  scale.
\newblock In \emph{ICLR}.

\bibitem[{He and McAuley(2016)}]{he2016ups}
Ruining He and Julian McAuley. 2016.
\newblock Ups and downs: Modeling the visual evolution of fashion trends with
  one-class collaborative filtering.
\newblock In \emph{WWW}, pages 507--517.

\bibitem[{Hermann et~al.(2015)Hermann, Kocisky, Grefenstette, Espeholt, Kay,
  Suleyman, and Blunsom}]{hermann2015teaching}
Karl~Moritz Hermann, Tomas Kocisky, Edward Grefenstette, Lasse Espeholt, Will
  Kay, Mustafa Suleyman, and Phil Blunsom. 2015.
\newblock Teaching machines to read and comprehend.
\newblock \emph{NIPS}, 28:1693--1701.

\bibitem[{Katharopoulos et~al.(2020)Katharopoulos, Vyas, Pappas, and
  Fleuret}]{katharopoulos2020transformers}
Angelos Katharopoulos, Apoorv Vyas, Nikolaos Pappas, and Fran{\c{c}}ois
  Fleuret. 2020.
\newblock Transformers are rnns: Fast autoregressive transformers with linear
  attention.
\newblock In \emph{ICML}, pages 5156--5165.

\bibitem[{Kitaev et~al.(2020)Kitaev, Kaiser, and Levskaya}]{Kitaev2020reformer}
Nikita Kitaev, Lukasz Kaiser, and Anselm Levskaya. 2020.
\newblock Reformer: The efficient transformer.
\newblock In \emph{ICLR}.

\bibitem[{Lan et~al.(2020)Lan, Chen, Goodman, Gimpel, Sharma, and
  Soricut}]{Lan2020albert}
Zhenzhong Lan, Mingda Chen, Sebastian Goodman, Kevin Gimpel, Piyush Sharma, and
  Radu Soricut. 2020.
\newblock {ALBERT:} {A} lite {BERT} for self-supervised learning of language
  representations.
\newblock In \emph{ICLR}.

\bibitem[{Parikh et~al.(2016)Parikh, T{\"a}ckstr{\"o}m, Das, and
  Uszkoreit}]{parikh2016decomposable}
Ankur Parikh, Oscar T{\"a}ckstr{\"o}m, Dipanjan Das, and Jakob Uszkoreit. 2016.
\newblock A decomposable attention model for natural language inference.
\newblock In \emph{EMNLP}, pages 2249--2255.

\bibitem[{Pennington et~al.(2014)Pennington, Socher, and
  Manning}]{pennington2014glove}
Jeffrey Pennington, Richard Socher, and Christopher Manning. 2014.
\newblock Glove: Global vectors for word representation.
\newblock In \emph{EMNLP}, pages 1532--1543.

\bibitem[{Radford et~al.(2019)Radford, Wu, Child, Luan, Amodei, and
  Sutskever}]{radford2019language}
Alec Radford, Jeffrey Wu, Rewon Child, David Luan, Dario Amodei, and Ilya
  Sutskever. 2019.
\newblock Language models are unsupervised multitask learners.
\newblock \emph{OpenAI blog}, 1(8):9.

\bibitem[{Tay et~al.(2021)Tay, Bahri, Metzler, Juan, Zhao, and
  Zheng}]{tay2021synthesizer}
Yi~Tay, Dara Bahri, Donald Metzler, Da-Cheng Juan, Zhe Zhao, and Che Zheng.
  2021.
\newblock Synthesizer: Rethinking self-attention for transformer models.
\newblock In \emph{ICML}, pages 10183--10192.

\bibitem[{Tay et~al.(2020)Tay, Dehghani, Bahri, and Metzler}]{tay2020efficient}
Yi~Tay, Mostafa Dehghani, Dara Bahri, and Donald Metzler. 2020.
\newblock Efficient transformers: A survey.
\newblock \emph{arXiv preprint arXiv:2009.06732}.

\bibitem[{Vaswani et~al.(2017)Vaswani, Shazeer, Parmar, Uszkoreit, Jones,
  Gomez, Kaiser, and Polosukhin}]{vaswani2017attention}
Ashish Vaswani, Noam Shazeer, Niki Parmar, Jakob Uszkoreit, Llion Jones,
  Aidan~N Gomez, {\L}ukasz Kaiser, and Illia Polosukhin. 2017.
\newblock Attention is all you need.
\newblock In \emph{NIPS}, pages 5998--6008.

\bibitem[{Wang et~al.(2020{\natexlab{a}})Wang, Wu, Liu, and Xie}]{wang2020fine}
Heyuan Wang, Fangzhao Wu, Zheng Liu, and Xing Xie. 2020{\natexlab{a}}.
\newblock Fine-grained interest matching for neural news recommendation.
\newblock In \emph{ACL}, pages 836--845.

\bibitem[{Wang et~al.(2020{\natexlab{b}})Wang, Li, Khabsa, Fang, and
  Ma}]{wang2020linformer}
Sinong Wang, Belinda Li, Madian Khabsa, Han Fang, and Hao Ma.
  2020{\natexlab{b}}.
\newblock Linformer: Self-attention with linear complexity.
\newblock \emph{arXiv preprint arXiv:2006.04768}.

\bibitem[{Wang et~al.(2017)Wang, He, Nie, and Chua}]{wang2017item}
Xiang Wang, Xiangnan He, Liqiang Nie, and Tat-Seng Chua. 2017.
\newblock Item silk road: Recommending items from information domains to social
  users.
\newblock In \emph{SIGIR}, pages 185--194.

\bibitem[{Wu et~al.(2019)Wu, Wu, Ge, Qi, Huang, and Xie}]{wu2019nrms}
Chuhan Wu, Fangzhao Wu, Suyu Ge, Tao Qi, Yongfeng Huang, and Xing Xie. 2019.
\newblock Neural news recommendation with multi-head self-attention.
\newblock In \emph{EMNLP}, pages 6390--6395.

\bibitem[{Wu et~al.(2021{\natexlab{a}})Wu, Wu, Qi, and Huang}]{wu2021plm}
Chuhan Wu, Fangzhao Wu, Tao Qi, and Yongfeng Huang. 2021{\natexlab{a}}.
\newblock Empowering news recommendation with pre-trained language models.
\newblock In \emph{SIGIR}, pages 1652--1656. {ACM}.

\bibitem[{Wu et~al.(2021{\natexlab{b}})Wu, Wu, Qi, and Huang}]{wu2021hi}
Chuhan Wu, Fangzhao Wu, Tao Qi, and Yongfeng Huang. 2021{\natexlab{b}}.
\newblock Hi-transformer: Hierarchical interactive transformer for efficient
  and effective long document modeling.
\newblock In \emph{ACL}.

\bibitem[{Wu et~al.(2020)Wu, Qiao, Chen, Wu, Qi, Lian, Liu, Xie, Gao, Wu
  et~al.}]{wu2020mind}
Fangzhao Wu, Ying Qiao, Jiun-Hung Chen, Chuhan Wu, Tao Qi, Jianxun Lian,
  Danyang Liu, Xing Xie, Jianfeng Gao, Winnie Wu, et~al. 2020.
\newblock Mind: A large-scale dataset for news recommendation.
\newblock In \emph{ACL}, pages 3597--3606.

\bibitem[{Zaheer et~al.(2020)Zaheer, Guruganesh, Dubey, Ainslie, Alberti,
  Onta{\~{n}}{\'{o}}n, Pham, Ravula, Wang, Yang, and Ahmed}]{zaheer2020big}
Manzil Zaheer, Guru Guruganesh, Kumar~Avinava Dubey, Joshua Ainslie, Chris
  Alberti, Santiago Onta{\~{n}}{\'{o}}n, Philip Pham, Anirudh Ravula, Qifan
  Wang, Li~Yang, and Amr Ahmed. 2020.
\newblock Big bird: Transformers for longer sequences.
\newblock In \emph{NeurIPS}.

\bibitem[{Zhang et~al.(2021)Zhang, Gong, Shen, Li, Lv, Duan, and
  Chen}]{zhang2021pooling}
Hang Zhang, Yeyun Gong, Yelong Shen, Weisheng Li, Jiancheng Lv, Nan Duan, and
  Weizhu Chen. 2021.
\newblock Poolingformer: Long document modeling with pooling attention.
\newblock In \emph{ICML}, volume 139, pages 12437--12446.

\end{thebibliography}

\appendix

\section{Appendix}

\subsection{Experimental Environment}

Our experiments are conducted on a cloud Linux server with Ubuntu 16.04 operating system.
The codes are written in Python 3.6 using the Keras library 2.2.4 with Tensorflow 1.12 backend.
The GPU type is Nvidia Tesla V100 with 32GB GPU memory.
Each experiment is run by a single thread.

\subsection{Preprocessing}

In our experiments, we use the NLTK tool to preprocess the texts.
We use the word\_tokenize and function to convert the input texts into token sequences.
The word embeddings of out-of-vocabulary words are filled with random vectors that have the same mean and co-variation values as other words.

\subsection{Hyperparameter Settings}

The detailed hyperparameter settings on each dataset used in this paper are listed in Table~\ref{hyper}.

\begin{table*}[h]
\resizebox{0.98\textwidth}{!}{
\begin{tabular}{lcccccc}
\hline
\multicolumn{1}{c}{Method}        & Amazon       & IMDB         & \begin{tabular}[c]{@{}c@{}}MIND\\ (classification)\end{tabular} & \begin{tabular}[c]{@{}c@{}}MIND\\ (recommendation)\end{tabular}              & CNN/DailyMail & PubMed       \\ \hline
\# Encoder Layer                  & 2            & 2            & 2                                                               & 1                                                                            & 4             & 2            \\
\# Decoder Layer                  & -            & -            & -                                                               & -                                                                            & 4             & 4            \\
\# Global Token                   & 8            & 8            & 16                                                              & 8                                                                            & 64            & 64           \\
\# Random Token                   & 8            & 8            & 16                                                              & 8                                                                            & 64            & 64           \\
Window size (Longformer, BigBird) & 8            & 8            & 16                                                              & 8                                                                            & 64            & 64           \\
Window size (Poolingformer)       & 64           & 64           & 256                                                             & 64                                                                           & 256           & 256          \\
Block length (BigBird)            & 4            & 4            & 8                                                               & 4                                                                            & 32            & 32           \\
Projection dimension (Linformer)  & 16           & 16           & 16                                                              & 16                                                                           & 64            & 64           \\
\# Attention Head                 & 16           & 16           & 16                                                              & 16                                                                           & 16            & 16           \\
Beam Size                         & -            & -            & -                                                               & -                                                                            & 5             & 5            \\
Maximum Text Length (Transformer)                        & 512            & 512            & 512                                                               & 512                                                                            & 512             & 512            \\
Maximum Text Length (others)                      & 512            & 1,024            & 2,048                                                               & 48                                                                            & 2,048             & 4,096            \\
Maximum User Behaviors                        & -            & -            & -                                                               & 50                                                                            & -             & -          \\
Hidden dimension                  & 256          & 256          & 256                                                             & 256                                                                          & 256           & 256          \\
Loss                              & Crossentropy & Crossentropy & Crossentropy                                                    & Crossentropy                                                                 & Crossentropy  & Crossentropy \\
Batch Size                        & 64           & 64           & 64                                                              & 64                                                                           & 32            & 32           \\
Optimizer                         & Adam         & Adam         & Adam                                                            & Adam                                                                         & Adam          & Adam         \\
Learning Rate                     & 1e-3         & 1e-3         & 1e-3                                                            & \begin{tabular}[c]{@{}c@{}}3e-6 for PLM-NR\\ 1e-4 for others\end{tabular} & 1e-4          & 1e-4         \\
Max Epochs                         & 3         & 3         & 3                                                            & 3                                                                         & 12          & 15         \\
Dropout                           &  0.2         & 0.2         & 0.2 & 0.2 & 0.2 & 0.2                                                                                                                                                                                              \\ \hline
\end{tabular}

}
\caption{Detailed hyperparameter settings on each dataset.}\label{hyper}
\end{table*}

\end{document}